%% file: emnlp2021.tex
\pdfoutput=1

\documentclass[11pt]{article}

\usepackage{emnlp2021}

\usepackage{times}
\usepackage{latexsym}

\usepackage[T1]{fontenc}

\usepackage[utf8]{inputenc}

\usepackage{microtype}
\usepackage{booktabs}
\usepackage{graphicx}
\usepackage{multirow}
\usepackage{amsmath, amssymb}
\usepackage{enumitem}

\DeclareMathOperator*{\argmax}{arg\,max}
\newcommand{\be}{\mathbf{e}}

\newcommand{\bv}{\mathbf{v}}
\newcommand{\bx}{\mathbf{x}}

\newcommand{\bz}{\mathbf{z}}
\newcommand{\bpi}{\boldsymbol\pi}
\newcommand{\calL}{\mathcal{L}}
\newcommand{\calV}{\mathcal{V}}
\newcommand{\calX}{\mathcal{X}}
\newcommand{\calY}{\mathcal{Y}}

\title{Gradient-based Adversarial Attacks against Text Transformers}

\author{
Chuan Guo$^{*}$ \qquad Alexandre Sablayrolles\thanks{$^{*}$Equal contribution.} \qquad Hervé Jégou \qquad Douwe Kiela\\
Facebook AI Research
}

\begin{document}
\maketitle
\begin{abstract}
%
%
We propose the first general-purpose gradient-based attack against transformer models. Instead of searching for a single adversarial example, we search for a distribution of adversarial examples parameterized by a continuous-valued matrix, hence enabling gradient-based optimization. We empirically demonstrate that our white-box attack attains state-of-the-art attack performance on a variety of natural language tasks. Furthermore, we show that a powerful black-box transfer attack, enabled by sampling from the adversarial distribution, matches or exceeds existing methods, while only requiring hard-label outputs.
\end{abstract}

\input{sections/intro}
\input{sections/background}
\input{sections/method}
\input{sections/experiment}
\input{sections/conclusion}

\bibliography{anthology,custom}
\bibliographystyle{acl_natbib}


\end{document}

%% file: sections/intro.tex
\section{Introduction}
\label{sec:intro}


Deep neural networks are sensitive to small, often imperceptible changes in the input, as evidenced by the existence of so-called \emph{adversarial examples}~\cite{biggio2013evasion, szegedy2013intriguing}.

The dominant method for constructing adversarial examples defines an \emph{adversarial loss}, which encourages prediction error, and then minimizes the adversarial loss with established optimization techniques. To ensure that the perturbation is hard to detect by humans, existing methods often introduce a perceptibility constraint into the optimization problem. Variants of this general strategy have been successfully applied to image and speech data~\cite{carlini2017towards, madry2017towards, carlini2018audio}.

However, optimization-based search strategies for obtaining adversarial examples are much more challenging with text data. Attacks against continuous data types such as image and speech utilize gradient descent for superior efficiency, but the discrete nature of natural languages prohibits such first-order techniques. In addition, perceptibility for continuous data can be approximated with $L_2$- and $L_\infty$-norms, but such metrics are not readily applicable to text data. To circumvent this issue, some existing attack approaches have opted for heuristic word replacement strategies and optimizing by greedy or beam search using black-box queries~\cite{jin2020bert, li2020contextualized, li-etal-2020-bert-attack, garg-ramakrishnan-2020-bae}. Such heuristic strategies typically introduce unnatural changes that are grammatically or semantically incorrect~\cite{morris-etal-2020-reevaluating}.

\begin{figure*}
    \centering
    \includegraphics[width=\textwidth]{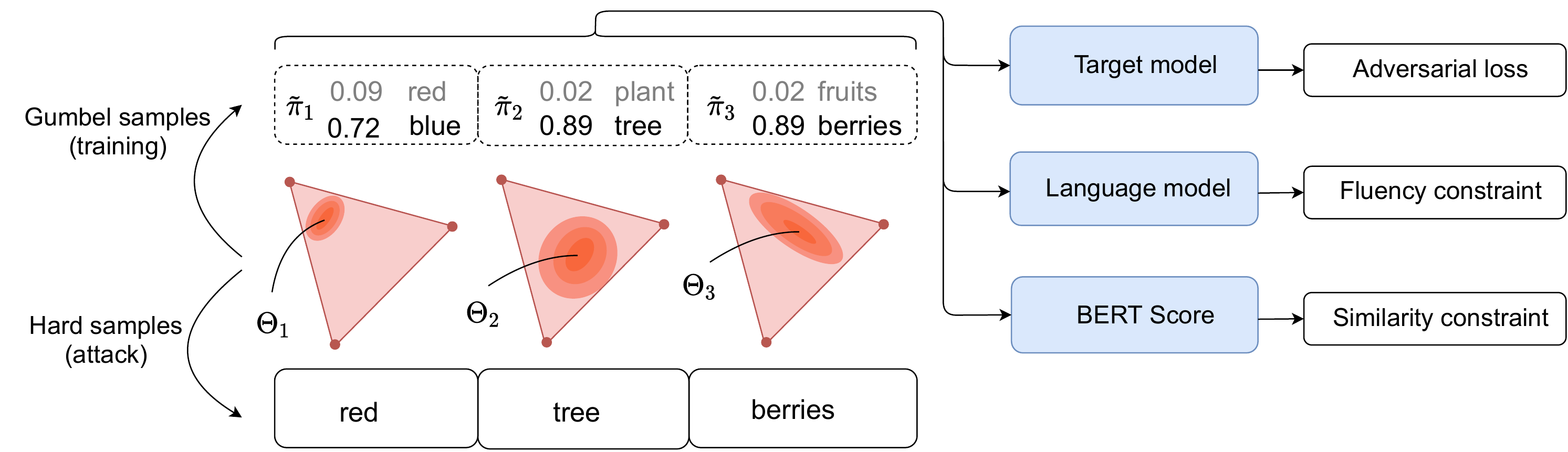}
    \caption{\label{fig:overview}
    Overview of our attack framework. The parameter matrix $\Theta$ is used to sample a sequence of probability vectors $\tilde{\pi}_1,\ldots,\tilde{\pi}_n$, which is forwarded through three (not necessarily distinct) models: (i) the target model for computing the adversarial loss, (ii) the language model for the fluency constraint, and (iii) the BERTScore model for the semantic similarity constraint. Due to the differentiable nature of each loss component and of the Gumbel-softmax distribution, our framework is fully differentiable, hence enabling gradient-based optimization. 
    }
\end{figure*}

In this paper, we propose a general-purpose framework for gradient-based adversarial attacks, and apply it against transformer models on text data.
Our framework, \textbf{GBDA} (Gradient-based Distributional Attack), consists of two key components that circumvent the difficulties of gradient descent for discrete data under perceptibility constraints. First, instead of constructing a single adversarial example, we search for an \emph{adversarial distribution}. We instantiate examples with the Gumbel-softmax distribution~\cite{jang2016categorical},
parameterized by a continuous matrix of coefficients that we optimize with a vanilla gradient-based method. Second, we enforce perceptibility and fluency using BERTScore~\cite{zhang2019bertscore} and language model perplexity, respectively, both of which are differentiable and can be added to the objective function as soft constraints. The combination of these two components enables powerful, efficient, gradient-based text adversarial attacks.

Our approach overcomes many ad-hoc constraints and limitations from the existing adversarial NLP literature. For instance, some black-box attacks~\cite{li-etal-2020-bert-attack} generate token-level candidates for replacement and then rescore all the combinations of these subword tokens to find the best word-level candidate: this can lead to exponential blow-up of the search space for rare or composite words. Another problem encountered by mask predictions for substitutions is that tokens have to be modified sequentially in arbitrary order.
In contrast, GBDA operates at the token level and the fluency of token combinations is handled directly by the language model constraint. 
This makes our method generic and potentially applicable to any model for text sequence prediction.

We empirically demonstrate the efficacy of GBDA against several transformer models. In addition, we also evaluate under the transfer-based black-box threat model by sampling from the optimized adversarial distribution and querying against a different, potentially unknown target model. On a variety of tasks including news/article categorization, sentiment analysis, and natural language inference, our method achieves state-of-the-art attack success rate, while preserving fluency, grammatical correctness, and a high level of semantic similarity to the original input. 

In summary, the main contributions of our paper are as follows:
\begin{enumerate}
    \itemsep -1ex
    \item We define a parameterized distribution of adversarial examples and optimize it using gradient-based methods. In contrast, most prior work construct a single adversarial example using black-box search.
    \item By incorporating differentiable fluency and semantic similarity constraints into the adversarial loss, our white-box attack produces more natural adversarial texts while setting a new state-of-the-art success rate.
    \item The adversarial distribution can be sampled efficiently to query different target models in a black-box setting. This enables a powerful transfer attack that matches or exceeds the performance of existing black-box attack. Compared to prior work that operate a continuous-valued output from the target model, this transfer attack only requires hard-label outputs.
\end{enumerate}

%% file: sections/background.tex
\section{Background}
\label{sec:background}

Adversarial examples constitute a class of robustness attacks against neural networks. Let $h : \calX \rightarrow \calY$ be a classifier where  $\calX, \calY$ are the input and output domains, respectively. Suppose that $\bx \in \calX$ is a test input that the model correctly predicts as the label $y = h(\bx) \in \calY$. An (untargeted) adversarial example is a sample $\bx' \in \calX$ such that $h(\bx') \neq y$ but $\bx'$ and $\bx$ are imperceptibly close.

The notion of perceptibility is introduced so that $\bx'$ preserves the semantic meaning of $\bx$ for a human observer. At a high level, $\bx'$ constitutes an attack on the model's robustness if a typical human would not misclassify $\bx'$ but the model $h$ does. For image data, since the input domain $\calX$ is a subset of the Euclidean space $\mathbb{R}^d$, a common surrogate for perceptibility is a distance metric such as the Euclidean distance or the Chebyshev distance. In general, one can define a perceptibility metric $\rho : \calX \times \calX \rightarrow \mathbb{R}_{\geq 0}$ and a threshold $\epsilon > 0$ so that $\bx'$ is considered imperceptible to $\bx$ if $\rho(\bx, \bx') \leq \epsilon$.

\paragraph{Search problem formulation.} The process of finding an adversarial example is typically modeled as an optimization problem. For classification, the model $h$ outputs a \emph{logit vector} $\phi_h(\bx) \in \mathbb{R}^K$ such that $y = \argmax_k \phi_h(\bx)_k$. To encourage the model to misclassify an input, one can define an \emph{adversarial loss} such as the margin loss:
\begin{align}
    \ell_\text{margin}(&\bx, y; h) = \nonumber \\
    &\max\left( \phi_h(\bx)_y - \max_{k \neq y} \phi_h(\bx)_k + \kappa, 0 \right),
    \label{eq:margin_loss}
\end{align}
so that the model misclassifies $\bx$ by a margin of $\kappa > 0$ when the loss is 0. The margin loss has been used in attack algorithms for image data~\cite{carlini2017towards}.

Given an adversarial loss $\ell$, the process of constructing an adversarial example can be cast as a constrained optimization problem:
\begin{align}
    \label{eq:adv_opt_hard}
    \min_{\bx' \in \mathcal{X}} \enspace &\ell(\bx', y; h) \quad \text{subject to} \; \rho(\bx, \bx') \leq \epsilon.
\end{align}
An alternative formulation is to relax the constraint into a soft constraint with $\lambda > 0$:
\begin{equation}
    \min_{\bx' \in \mathcal{X}} \; \ell(\bx', y; h) + \lambda \cdot \rho(\bx, \bx'),
\end{equation}
which can then be solved using gradient-based optimizers if the constraint function $\rho$ is differentiable.

\subsection{Text Adversarial Examples}

Although the search problem formulation in \autoref{eq:adv_opt_hard} has been widely applied to continuous data such as image and speech, it does not directly apply to text data because (1) the data space $\calX$ is discrete, hence not permitting gradient-based optimization; and (2) the constraint function $\rho$ is difficult to define for text data. In fact, both issues arise when considering attacks against \emph{any} discrete input domain, but the latter is especially relevant for text data due to the sensitivity of natural language. For instance, inserting the word \emph{not} into a sentence can negate the meaning of the whole sentence despite having a token-level edit distance of 1.

\paragraph{Prior work.} Several attack algorithms have been proposed to circumvent these two issues, using a multitude of approaches. For attacks that operate on the character level, perceptibility can be approximated by the number of character edits, \emph{i.e.}, replacements, swaps, insertions and deletions~\cite{ebrahimi2017hotflip, li2018textbugger, gao2018black}. Attacks that operate on the word level adopt heuristics such as synonym substitution~\cite{samanta2017towards, zang-etal-2020-word} or replacing words by ones with similar word embeddings~\cite{alzantot2018generating, ren-etal-2019-generating, jin2020bert}. More recent attacks have also leveraged masked language models such as BERT~\cite{devlin-etal-2019-bert} to generate word substitutions by replacing masked tokens~\cite{garg-ramakrishnan-2020-bae, li2020contextualized, li-etal-2020-bert-attack}. Most of the aforementioned attacks follow the common recipe of proposing character-level or word-level perturbations to generate a constrained candidate set and optimizing the adversarial loss greedily or using beam search.

\paragraph{Shortcomings in prior work.} Despite the plethora of attacks against natural language models, their efficacy remains subpar compared to attacks against other data modalities. Both character-level and word-level changes are still relatively detectable, especially as such changes often introduce misspellings, grammatical errors, and other artifacts of unnaturalness in the perturbed text~\cite{morris-etal-2020-reevaluating}. Moreover, prior attacks mostly query the target model $h$ as a black-box and rely on zeroth-order strategies for minimizing the adversarial loss, resulting in sub-optimal performance.

For instance, BERT-Attack~\cite{li-etal-2020-bert-attack}---arguably the state-of-the-art attack against BERT---only reduces the test accuracy of the target model on the AG News dataset~\cite{zhang2015character} from $94.2$ to $10.6$. In comparison, attacks against image models can consistently reduce the model's accuracy to 0 on almost all computer vision tasks~\cite{akhtar2018threat}. This gap in performance raises the question of whether gradient-based search can produce more fluent and optimal adversarial examples on text data. In this work, we show that our gradient-based attack can reduce the same model's accuracy from $94.2$ to $2.5$ while being \emph{more semantically-faithful} to the original text.

\subsection{Other Attacks}

While most work on adversarial text attack falls within the formulation that we defined in \autoref{sec:background}, other notions of text adversarial attacks exist as well. One class of such attacks is known as a universal adversarial trigger---a short snippet of text that when appended to any input, causes the model to misclassify~\cite{wallace2019universal, song2020universal}. However, such triggers often contain unnatural combinations of words or tokens, and hence are very perceptible to a human observer.

Our work falls within the general area of adversarial learning, and many prior works in this area have explored the notion of adversarial example on different data modalities. While the most prominent data modality by far is image, adversarial examples can be constructed for speech~\cite{carlini2018audio} and graphs~\cite{dai2018adversarial, zugner2018adversarial} as well.

%% file: sections/method.tex
\section{GBDA: Gradient-based Distributional Attack}
\label{sec:method}

In this section, we detail GBDA---our general-purpose framework for gradient-based text attacks against transformers. Our framework leverages two important insights: (1) we define a parameterized \emph{adversarial distribution} that enables gradient-based search using the Gumbel-softmax~\cite{jang2016categorical}; and (2) we promote fluency and semantic faithfulness of the perturbed text using soft constraints on both perplexity and semantic similarity.

\subsection{Adversarial Distribution}
\label{sec:adv_dist}

Let $\bz = z_1 z_2 \cdots z_n$ be a sequence of tokens where each $z_i \in \calV$ is a token from a fixed vocabulary $\calV = \{1,\ldots,V\}$. Consider a distribution $P_\Theta$ parameterized by a matrix $\Theta \in \mathbb{R}^{n \times V}$, which draws samples $\bz \sim P_\Theta$ by independently sampling each token
\begin{equation}
    \label{eq:categorical}
    z_i \sim \text{Categorical}(\pi_i),
\end{equation}
where $\pi_i = \text{Softmax}(\Theta_i)$ is a vector of token probabilities for the $i$-th token.

We aim to optimize the parameter matrix $\Theta$ so that samples $\bz \sim P_\Theta$ are adversarial examples for the model $h$. To do so, we define the objective function for this goal as:
\begin{equation}
    \label{eq:obj_func}
    \min_{\Theta \in \mathbb{R}^{n \times V}} \mathbb{E}_{\bz \sim P_\Theta} \ell(\bz, y; h),
\end{equation}
where $\ell$ is a chosen adversarial loss.

\paragraph{Extension to probability vector inputs.} The objective function in \autoref{eq:obj_func} is non-differentiable due to the discrete nature of the categorical distribution. Instead, we propose a relaxation of \autoref{eq:obj_func} by first extending the model $h$ to take probability vectors as input, and then use the Gumbel-softmax approximation~\cite{jang2016categorical} of the categorical distribution to derive the gradient.

Transformer models take as input a sequence of tokens that are converted to embedding vectors using 
a lookup table. 
In particular, let $\be(\cdot)$ be the embedding function so that the input embedding for the token $z_i$ is $\be(z_i) \in \mathbb{R}^d$ for some embedding dimension $d$. Given a probability vector $\pi_i$ that specifies the sampling probability of the token $z_i$, we define
\begin{equation}
    \label{eq:vector_input}
    \be(\pi_i) = \sum_{j=1}^V (\pi_i)_j \be(j)
\end{equation}
as the embedding vector corresponding to the probability vector $\pi_i$. Note that if $\pi_i$ is a one-hot vector corresponding to the token $z_i$ then $\be(\pi_i) = \be(z_i)$. We extend the notation for an input sequence of probability vectors $\bpi = \pi_1 \cdots \pi_n$ as:
$$\be(\bpi) = \be(\pi_1) \cdots \be(\pi_n)$$
by concatenating the input embeddings.


\paragraph{Computing gradients using Gumbel-softmax.} Extending the model $h$ to take probability vectors as input allows us to leverage the Gumbel-softmax approximation to derive smooth estimates of the gradient of \autoref{eq:obj_func}. Samples $\tilde{\bpi} = \tilde{\pi}_1 \cdots \tilde{\pi}_n$ from the Gumbel-softmax distribution $\tilde{P}_\Theta$ are drawn according to the process:
\begin{equation}
    \label{eq:gumbel_softmax}
    (\tilde{\pi}_i)_j := \frac{\exp((\Theta_{i,j} + g_{i,j}) / T)}{\sum_{v=1}^{V} \exp((\Theta_{i,v} + g_{i,v}) / T)},
\end{equation}
where $g_{i,j} \sim \text{Gumbel}(0,1)$ and $T > 0$ is a temperature parameter that controls the smoothness of the Gumbel-softmax distribution. 
As $T \rightarrow 0$, this distribution converges towards the categorical distribution in \autoref{eq:categorical}.

We can now optimize $\Theta$ using gradient descent by defining a smooth approximation of the objective function in \autoref{eq:obj_func}:
\begin{equation}
    \label{eq:smooth_obj}
    \min_{\Theta \in \mathbb{R}^{n \times V}} \mathbb{E}_{\tilde{\bpi} \sim \tilde{P}_\Theta} \ell(\be(\tilde{\bpi}), y; h),
\end{equation}
The expectation can be estimated using stochastic samples of $\tilde{\bpi} \sim \tilde{P}_\Theta$.

\subsection{Soft Constraints}
\label{sec:soft_constraints}

Black-box attacks based on heuristic replacements can only constrain the perturbation by proposing changes that fall within the constraint set, \emph{e.g.}, limiting edit distance, replacing words by ones with high word embedding similarity, etc. In contrast, our adversarial distribution formulation can readily incorporate any differentiable constraint function as a part of the objective. We leverage this advantage to include both fluency and semantic similarity constraints in order to produce more fluent and semantically-faithful adversarial texts.

\paragraph{Fluency constraint with a language model.}
Causal language models (CLMs) are trained with the objective of next token prediction by maximizing the likelihood given previous tokens. This allows the computation of likelihoods for any sequence of tokens. More specifically, given a CLM $g$ with log-probability outputs, the negative log-likelihood (NLL) of a sequence $\bx = x_1 \cdots x_n$ is given autoregressively by:
\begin{equation*}
    \mathrm{NLL}_g(\bx) = -\sum_{i=1}^n \log p_g(x_i \mid x_1 \cdots x_{i-1}),
\end{equation*}
where $\log p_g(x_i \mid x_1 \cdots x_{i-1}) = g(x_1 \cdots x_{i-1})_{x_i}$ is the cross-entropy between the delta distribution on token $x_i$ and the predicted token distribution $g(x_1 \cdots x_{i-1})$ for $i=1,\ldots,n$.

In our adversarial distribution formulation, since the inputs are vectors of token probabilities, we extend the definition of NLL to:
\begin{equation*}
    \mathrm{NLL}_g(\bpi) := -\sum_{i=1}^n \log p_g(\pi_i \mid \pi_1 \cdots \pi_{i-1}),
\end{equation*}
with
\begin{align*}
    -\log p_g(&\pi_i \mid \pi_1 \cdots \pi_{i-1}) = \\
    &-\sum_{j=1}^V (\pi_i)_j g(\be(\pi_1) \cdots \be(\pi_{i-1}))_j
\end{align*}
being the cross-entropy between the next token distribution $\pi_i$ and the predicted next token distribution $g(\be(\pi_1) \cdots \be(\pi_{i-1}))$. This extension coincides with the NLL for a token sequence $\bx$ when each $\pi_i$ is a delta distribution for the token $x_i$.

\paragraph{Similarity constraint with BERTScore.} Prior work on word-level attacks often used context-free embeddings such as word2vec~\cite{mikolov2013distributed} and GloVe~\cite{pennington2014glove} or synonym substitution to constrain semantic similarity between the original and perturbed text~\cite{alzantot2018generating, ren-etal-2019-generating, jin2020bert}. These constraints tend to produce out-of-context and unnatural changes that alter the semantic meaning of the perturbed text~\cite{garg-ramakrishnan-2020-bae}. Instead, we propose to use BERTScore~\cite{zhang2019bertscore}, a similarity score for evaluating text generation that captures the semantic similarity between pairwise tokens in contextualized embeddings of a transformer model.

Let $\bx = x_1 \cdots x_n$ and $\bx' = x'_1 \cdots x'_m$ be two token sequences and let $g$ be a language model that produces contextualized embeddings $\phi(\bx) = (\bv_1, \ldots, \bv_n)$ and $\phi(\bx') = (\bv'_1, \ldots, \bv'_m)$. The (recall) BERTScore between $\bx$ and $\bx'$ is defined as:
\begin{equation}
    \label{eq:bertscore}
    R_\mathrm{BERT}(\bx, \bx') = \sum_{i=1}^n w_i \max_{j=1,\ldots,m} \bv_i^\top \bv'_j,
\end{equation}
where $w_i := \mathrm{idf}(x_i) / \sum_{i=1}^n \mathrm{idf}(x_i)$ is the normalized inverse document frequency of the token $x_i$ computed across a corpus of data. We can readily substitute $\bx'$ with a sequence of probability vectors $\bpi = \pi_1 \cdots \pi_m$ as described in \autoref{eq:vector_input} and use $\rho_g(\bx, \bpi) = 1 - R_\mathrm{BERT}(\bx, \bpi)$ as a differentiable soft constraint.

\paragraph{Objective function.} We combine all the components in the previous sections into a final objective for gradient-based optimization. Our objective function uses the margin loss (\emph{cf.} \autoref{eq:margin_loss}) as the adversarial loss, and integrates the fluency constraint with a causal language model $g$ and the BERTScore similarity constraint using contextualized embeddings of $g$:
\begin{align}
    \calL(\Theta) = &\mathbb{E}_{\tilde{\bpi} \sim \tilde{P}_\Theta} \ell(\be(\tilde{\bpi}), y; h) \nonumber \\ 
    &+ \lambda_{\mathrm{lm}} \: \mathrm{NLL}_g(\tilde{\bpi}) + \lambda_{\mathrm{sim}} \: \rho_g(\bx, \tilde{\bpi}),
\end{align}
where $\lambda_{\mathrm{lm}}, \lambda_{\mathrm{sim}} > 0$ are hyperparameters that control the strength of the soft constraints. We minimize $\calL(\Theta)$ stochastically using Adam~\cite{kingma2014adam} by sampling a batch of inputs from $\tilde{P}_\Theta$ at every iteration.

\begin{table*}[t!]
\resizebox{\textwidth}{!}{
\begin{tabular}{l|ccc|ccc|ccc}
\toprule
& \multicolumn{3}{c|}{\textbf{GPT-2}} & \multicolumn{3}{c|}{\textbf{XLM} (en-de)} & \multicolumn{3}{c}{\textbf{BERT}} \\
\textbf{Task} & \textbf{Clean Acc.} & \textbf{Adv. Acc.} & \textbf{Cosine Sim.} & \textbf{Clean Acc.} & \textbf{Adv. Acc.} & \textbf{Cosine Sim.} & \textbf{Clean Acc.} & \textbf{Adv. Acc.} & \textbf{Cosine Sim.} \\
\midrule
\textbf{DBPedia}                 & 99.2                & 5.2                & 0.91       & 99.1                & 7.6               & 0.80           & 99.2 & 7.1 & 0.80 \\
\textbf{AG News}                 & 94.8                & 6.6               & 0.90      & 94.4                & 5.4                & 0.87           & 95.1 & 2.5 & 0.82 \\
\textbf{Yelp}                 & 97.8                &   2.9             & 0.94         & 96.3                & 3.4                & 0.93                 & 97.3 & 4.7 & 0.92 \\
\textbf{IMDB}                 & 93.8                & 7.6                & 0.98        & 87.6                & 0.1                & 0.97                 & 93.0 & 3.0 & 0.92 \\
\textbf{MNLI} (m.)                 & 81.7                    & 2.8/11.0                   & 0.82/0.88           & 76.9                    & 1.3/8.4                   & 0.74/0.80                    & 84.6 & 7.1/10.2 & 0.87/0.92 \\
\textbf{MNLI} (mm.)                 & 82.5                    & 4.2/13.5                   & 0.85/0.88           & 76.3                    & 1.3/8.9                   & 0.75/0.80                    & 84.5 & 7.4/8.8 & 0.89/0.93 \\  
\bottomrule
\end{tabular}
}
\caption{Result of white-box attack against three transformer models: GPT-2, XLM (en-de), and BERT. Our attack is able to reduce the target model's accuracy to below $10\%$ in almost all cases, while maintaining a high level of semantic similarity (cosine similarity of higher than $0.8$ using USE embeddings).}
\label{tab:eval_whitebox}
\end{table*}

\subsection{Sampling Adversarial Texts}

Once $\Theta$ has been optimized, we can sample from the adversarial distribution $P_\Theta$ to construct adversarial examples. Since the loss function $\calL(\Theta)$ that we optimize is an approximation of the objective in \autoref{eq:obj_func}, it is possible that some samples are not adversarial even when $\calL(\Theta)$ is successfully minimized. Hence, in practice, we draw multiple samples $\bz \sim P_\Theta$ and stop sampling either when the model misclassifies the sample or when we reach a maximum number of samples.

Note that this stage could technically allow us to add hard constraints to the examples we generate, \emph{e.g.}, manually filter out adversarial examples that do not seem natural. 
In our case, we do not add any extra hard constraint and only verify that the generated example is misclassified by the model.

\paragraph{Transfer to other models.} Since drawing from the distribution $P_\Theta$ could potentially generate an infinite stream of adversarial examples, we can leverage these generated samples to query a \emph{target model} that is different from $h$. This constitutes a black-box \emph{transfer attack} from the source model $h$. Moreover, our transfer attack does not require the target model to output continuous-valued scores, which most existing black-box attacks against transformers rely on~\cite{jin2020bert, garg-ramakrishnan-2020-bae, li2020contextualized, li-etal-2020-bert-attack}. We demonstrate in \autoref{sec:quantitative} that this transfer attack enabled by the adversarial distribution $P_\Theta$ is very effective at attacking a variety of target models.

%% file: sections/experiment.tex
\section{Experiments}
\label{sec:experiment}

In this section, we empirically validate our attack framework on a variety of natural language tasks. Code to reproduce our results is open sourced on GitHub\footnote{\url{https://github.com/facebookresearch/text-adversarial-attack}}.

\begin{table*}[t!]
    \centering
    \resizebox{\linewidth}{!}{
    \begin{tabular}{l|l|l}
\textbf{Attack} & \textbf{Prediction} & \textbf{Text} \\
\midrule
Original    &   Entailment  (83\%)  &   He found himself thinking in circles of worry and pulled himself back to his problem. \\ && He got lost in loops of worry, but snapped himself back to his problem. \\
\textcolor{blue}{GBDA} &   \textcolor{blue}{Neutral (95\%)}      &   He found himself thinking in circles of worry and pulled himself back to his problem. \\ && He got lost in loops of \textcolor{blue}{hell}, but snapped himself back to his problem. \\[0.5em]

Original    &   Contradiction (95\%)    &   You're the Desert Ghost. You're a living desert camel. \\
\textcolor{blue}{GBDA} &   \textcolor{blue}{Entailment (51\%)}       &   You're the Desert Ghost. You're a living desert \textcolor{blue}{animal}. \\[0.5em]

Original    &   Contradiction (98\%)    &   Pesticide concentrations should not exceed USEPA's Ambient Water Quality chronic criteria values where available. \\&& There is no assigned value for maximum pesticide concentration in water. \\
\textcolor{blue}{GBDA} &   \textcolor{blue}{Entailment (86\%)}       &   Pesticide concentrations should not exceed USEPA's Ambient Water Quality chronic criteria values where available. \\&& There is \textcolor{blue}{varying} assigned value for maximum pesticide concentration in water.

 

    \end{tabular}
    }
    \caption{\label{tab:examples}Examples of successful adversarial texts on the MNLI dataset.
    }
\end{table*}

\subsection{Setup}

\paragraph{Tasks.} We evaluate on several benchmark text classification datasets, including \textbf{DBPedia}~\citep{zhang2015character} and \textbf{AG News}~\cite{zhang2015character} for article/news categorization, \textbf{Yelp Reviews}~\cite{zhang2015character} and \textbf{IMDB}~\cite{maas2011learning} for binary sentiment classification, and \textbf{MNLI}~\cite{williams2017broad} for natural language inference. The MNLI dataset contains two evaluation sets: matched (m.) and mismatched (mm.), 
corresponding to whether the test domain is matched or mismatched with the training distribution.

\paragraph{Models.} We attack three transformer architectures with our gradient-based white-box attack: GPT-2~\cite{radford2019language}, XLM~\cite{lample2019cross} (using the en-de cross-lingual model), and BERT~\cite{devlin-etal-2019-bert}. For BERT, we use finetuned models from TextAttack~\cite{morris2020textattack} for all tasks except for DBPedia, where finetuned models are unavailable. For BERT on DBPedia and GPT-2/XLM on all tasks, we finetune a pretrained model to serve as the target model. 

The soft constraints described in \autoref{sec:soft_constraints} utilizes a CLM $g$ with the same tokenizer as the target model. For GPT-2 we use the pre-trained GPT-2 model without finetuning as $g$, and for XLM we use the checkpoint obtained after finetuning using the CLM objective. For masked language models such as BERT~\cite{devlin-etal-2019-bert}, we train a causal language model $g$ on WikiText-103 using the same tokenizer as the target model.

\paragraph{Baselines.} We compare against several recent attacks on text transformers: TextFooler~\cite{jin2020bert}, BAE~\cite{garg-ramakrishnan-2020-bae}, and BERT-Attack~\cite{li-etal-2020-bert-attack}. All baseline attacks are evaluated on finetuned BERT models from the TextAttack library. Since these attacks only require black-box query access to the target model, we evaluate both our white-box attack on the finetuned BERT model, as well as transfer attack from the GPT-2 model for fair comparison. Please refer to \autoref{sec:quantitative} for details of both attack settings.

\paragraph{Hyperparameters.} 
Our adversarial distribution parameter $\Theta$ is optimized using Adam~\cite{kingma2014adam} with a learning rate of $0.3$ and a batch size of $10$ for $100$ iterations. 
The distribution parameters $ \Theta $ are initialized to zero except $ \Theta_{i, j} = C $ where $x_i = j$ is the $i$-th token of the clean input. 
In practice we take $C \in {12, 15}$. 
We use $ \lambda_{\mathrm{perp}} = 1$ and cross-validate $\lambda_{\mathrm{sim}} \in [20, 200]$ and $\kappa \in \{3, 5, 10\}$ using held-out data. 

\subsection{Quantitative Evaluation}
\label{sec:quantitative}

\paragraph{White-box attacks.} We first evaluate the attack performance under the white-box setting. \autoref{tab:eval_whitebox} shows the result of our attacks against GPT-2, XLM (en-de), and BERT on different benchmark datasets. Following prior work~\cite{jin2020bert}, for each task, we randomly select $1000$ inputs from the task's test set as attack targets. After optimizing $\Theta$, we draw up to $100$ samples $\bz \sim P_\Theta$ until the model misclassifies $\bz$. The model's accuracy after attack (under the column ``Adv. Acc.'') is the accuracy evaluated on the last of the drawn samples.

Overall, our attack is able to successfully generate adversarial examples against all three models across the five benchmark datasets. The test accuracy can be reduced to below $10\%$ for almost all models and tasks. Following prior work, we also evaluate the semantic similarity between the adversarial example and the original input using the cosine similarity of Universal Sentence Encoders~\cite{cer2018universal} (USE). Our attack is able to consistently maintain a high level of semantic similarity to the original input in most cases (cosine similarity higher than $0.8$).

\paragraph{Transfer attacks.} We evaluate our attack against prior work under the black-box transfer attack setting. More specifically, for each model and task, we randomly select $1000$ target test inputs and optimize the adversarial distribution $P_\Theta$ on GPT-2. After optimizing $\Theta$, we draw up to $1000$ samples $\bz \sim P_\Theta$ and evaluate them on the target BERT model from the TextAttack~\cite{morris2020textattack} library until the model misclassifies $\bz$. This attack setting is strictly more restrictive than prior work because our query procedure only requires the target model to output a discrete label in order to decide when to stop sampling from $P_\Theta$, whereas prior work relied on a continuous-valued output score such as class probabilities.

\autoref{tab:eval_bert} shows the performance of our attack when transferred to finetuned BERT text classifiers. In almost all settings, GBDA is able to reduce the target model's accuracy to close to that of BERT-Attack or better within fewer number of queries. Moreover, the cosine similarity between the original input and the adversarial example is higher than that of BERT-Attack.

\begin{table}[t!]
\resizebox{\columnwidth}{!}{
\begin{tabular}{@{\ }l@{\ }c@{\ \ }c@{\ \ }c@{\ \ }c@{\ \ }c@{\ }}
\toprule
\textbf{Task} & \textbf{Clean Acc.} & \textbf{Attack Alg.} & \textbf{Adv. Acc.} & \textbf{\# Queries} & \textbf{Cosine Sim.} \\
\midrule
\multirow{3}{*}{\textbf{AG News}}   &\multirow{3}{*}{95.1}  &GBDA (ours)    & \bf{8.8}  & {\bf 107} & 0.69    \\
&                                                           &BERT-Attack    &10.6       & 213       & 0.63          \\
&                                                           &BAE            &13.0       & 419       & {\bf 0.75} \\
&                                                           &TextFooler     &12.6       & 357       & 0.57          \\
\midrule
\multirow{3}{*}{\textbf{Yelp}}  &\multirow{3}{*}{97.3}  &GBDA (ours)    & {\bf 2.6} & {\bf 43}  & 0.83  \\
&                                                       &BERT-Attack    & 5.1       & 273       & 0.77  \\
&                                                       &BAE            & 12.0      & 434       & {\bf 0.90} \\
&                                                       &TextFooler     & 6.6       & 743       & 0.74  \\
\midrule
\multirow{3}{*}{\textbf{IMDB}}  &\multirow{3}{*}{93.0}  &GBDA (ours)    & {\bf 8.5}    & {\bf 116} & 0.92  \\
&                                                       &BERT-Attack    & 11.4          & 454       & 0.86       \\
&                                                       &BAE            & 24.0          & 592       & {\bf 0.95} \\
&                                                       &TextFooler     & 13.6          & 1134      & 0.86        \\
\midrule
\multirow{3}{*}{\textbf{MNLI} (m.)}            & \multirow{3}{*}{84.6}   &  GBDA (ours)               & {\bf 2.3}/{\bf 10.8}               & 37/133 & 0.75/0.79                 \\
& & BERT-Attack  & 7.9/11.9               & {\bf 19}/{\bf 44} & 0.55/0.68                 \\
& & BAE         & 25.4/36.2             & 68/120    & \bf{0.88}/\bf{0.88} \\
& & TextFooler  & 9.6/25.3               & 78/152 & 0.57/0.65                 \\
\midrule
\multirow{3}{*}{\textbf{MNLI} (mm.)}            & \multirow{3}{*}{84.5}   &  GBDA (ours)              & {\bf 1.8}/{\bf 13.4}               & 30/159 &  0.76/0.80                 \\
& & BERT-Attack  & 7/13.7               & {\bf 24}/{\bf 43} & 0.53/0.69                 \\
& & BAE         & 19.2/30.3          &  75/110   & \bf{0.88}/\bf{0.88} \\
& & TextFooler  & 8.3/22.9               & 86/162 & 0.58/0.65                 \\
\bottomrule
\end{tabular}
}
\caption{Evaluation of black-box transfer attack from GPT-2 to finetuned BERT classifiers. Our attack is able exceed the attack performance of BERT-Attack and BAE, while maintaining a higher semantic similarity with fewer number of queries in most cases. Furthermore, our transfer attack does not require continuous-valued outputs, which all the baseline methods rely on.}
\label{tab:eval_bert}
\end{table}

\begin{figure*}
  \begin{minipage}{0.59\textwidth}
        \resizebox{\textwidth}{!}{
        \begin{tabular}{@{\ }l@{\ }c@{\ }c@{\ \ }c@{\ \ }c@{\ \ }c@{\ }}
        \toprule
        \textbf{Target Model}              & \textbf{Task} & \textbf{Clean\, Acc.} & \textbf{Adv.\, Acc.} & \textbf{\# Queries} & \textbf{Cosine Sim.} \\
        \midrule
        \multirow{3}{*}{\textbf{ALBERT}} & \textbf{AG News}             & 94.7                & 7.5                & 84 & 0.68                 \\
        & \textbf{Yelp}             & 97.5                & 5.9                & 76 & 0.79                 \\
        & \textbf{IMDB}             & 93.8                & 13.1               & 157 & 0.87                 \\
        \midrule
        \multirow{4}{*}{\textbf{RoBERTA}}       & \textbf{AG News}              & 94.7                & 10.7                & 130 & 0.67                 \\
        & \textbf{IMDB}             & 95.2                & 17.4                & 205 & 0.87                 \\
        & \textbf{MNLI} (m.)             & 88.1                & 4.1/15.1                & 63/179 & 0.69/0.76                 \\
        & \textbf{MNLI} (mm.)             & 87.8                & 3.2/15.9                & 51/189 & 0.69/0.78                 \\
        \midrule
        \multirow{3}{*}{\textbf{XLNet}}       & \textbf{IMDB}      & 93.8                & 12.1                & 149 & 0.87                 \\
        & \textbf{MNLI} (m.)             &  87.2               &  3.9/13.7               &  56/162 & 0.70/0.77                 \\
        & \textbf{MNLI} (mm.)             &  86.8               &  1.7/14.4               &  32/171 &  0.70/0.78                \\
        \bottomrule
        \end{tabular}
        }
        \captionof{table}{Result of black-box transfer attack from GPT-2 to other transformer models. Our attack is achieved by sampling from the same adversarial distribution $P_\Theta$ and is able to generalize to the three target transformer models considered in this study.}
        \label{tab:eval_blackbox}
  \end{minipage}
  \hfill
  \begin{minipage}{0.39\textwidth}
    \centering
    \includegraphics[width=\textwidth]{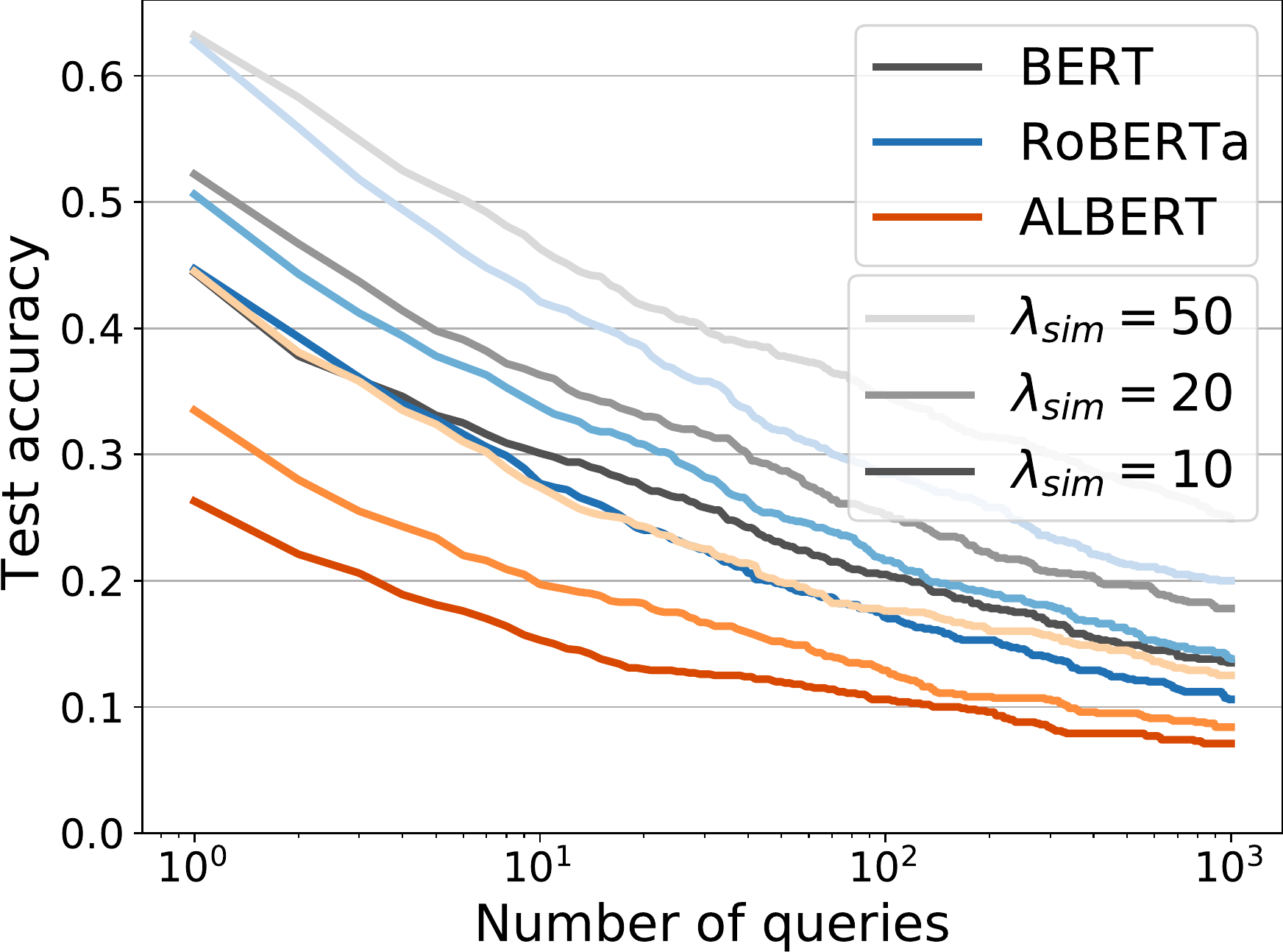}
    \captionof{figure}{
    \label{fig:lam_sim_transfer}
    Effect of the parameter $\lambda_\mathrm{sim}$ on transfer attack success rate. Lower $\lambda_\mathrm{sim}$ produces more aggressive changes, but also generalizes better to different target models.}
    \end{minipage}
\end{figure*}

We further evaluate our transfer attack against three finetuned transformer models from the TextAttack library: ALBERT~\cite{lan2019albert}, RoBERTa~\cite{liu2019roberta}, and XLNet~\cite{yang2019xlnet}. For this experiment, we use the same $\Theta$ optimized on GPT-2 for each of the target models. \autoref{tab:eval_blackbox} reports the performance of our attack after randomly sampling up to $1000$ times from $P_\Theta$. The attack performance is comparable to that of the transfer attack against BERT in \autoref{tab:eval_bert}, which means our adversarial distribution $P_\Theta$ is able to capture the common failure modes of a wide variety of transformer models. This result opens a practical avenue of attack against real world systems as the attacker requires very limited access to the target model in order to succeed.

\subsection{Analysis}

\paragraph{Sample adversarial texts.} Table~\ref{tab:examples} shows examples of our adversarial attack on text. 
Our method introduces minimal changes to the text, preserving most of the original sentence's meaning. 
Despite not explicitly constraining replaced words to have the same Part-Of-Speech tag, we observe that our soft penalties make the adversarial examples obey this constraint.
For instance, in the first example of Table~\ref{tab:examples}, "camel" is replaced with "animal" and "no" with "varying".

\paragraph{Effect of $\lambda_{\mathrm{sim}}$.}
Figure \ref{fig:lam_sim_transfer} shows the impact of the similarity constraint on transfer attack adversarial accuracy for GPT-2 on AG News. 
Each color corresponds to a different target model, whereas the color shade (from light to dark) indicates the value of the constraint hyperparameter: $\lambda_\mathrm{sim} = 50,20,10$.
A higher value of $\lambda_\mathrm{sim}$ reduces the aggressiveness of the perturbation, but also increases the number of queries required to achieve a given target adversarial accuracy.
This plot shows that there exists a trade-off between the similarity constraint, attack success rate, and query budget. 

\paragraph{Impact of the fluency constraint.}
\autoref{tab:fluency_constraint} shows adversarial examples for GPT-2 on AG News, generated with and without the fluency constraint. 
We fix all hyperparameters except for the fluency regularization constant $\lambda_\mathrm{lm}$, and sample successful adversarial texts from $P_\Theta$ after $\Theta$ has been optimized. It is evident that the fluency constraint promotes token combinations to form valid words and ensure grammatical correctness of the adversarial text. Without the fluency constraint, the adversarial text tends to contain nonsensical words.

\paragraph{Tokenization artifacts.}
Our attack operates entirely on tokenized inputs.
However, the input to the classification system is often in raw text form, which is then tokenized before being fed to the model.
Thus it is possible that we generate an adversarial example that, when converted to raw text, is not re-tokenized to the same set of tokens.

Consider this example: our adversarial example contains the tokens "jui-" and "cy", which decodes into "juicy", and is then re-encoded to "juic-" and "y". 
In practice, we observe that these re-tokenization artifacts are rare: the "token error rate" is around $2\%$. 
Furthermore, they do not impact adversarial accuracy by much: the re-tokenized example is in fact still adversarial. One potential mitigation strategy is to re-sample from $P_\Theta$ until the sampled text is stable under re-tokenization. 
Note that all our adversarial accuracy results are computed after re-tokenization.

\paragraph{Runtime.}
Our method relies on white-box optimization and thus necessitates forward and backward passes through the attacked model, the language model and the similarity model, which increases the per-query time compared to black-box attacks that only compute forward passes. 
However, this is compensated by a much more efficient optimization which brings the total runtime to $20s$ per generated example, on par with black-box attacks such as BERT-Attack~\cite{li-etal-2020-bert-attack}. 


\begin{table*}[t!]
    \centering
    \resizebox{\linewidth}{!}{
    \begin{tabular}{l|l|l}
\textbf{Attack} & \textbf{Prediction} & \textbf{Text} \\
\midrule
Original    & World (99\%) & Turkey a step closer to Brussels The European Commission is set to give the green light later today \\ & & to accession talks with Turkey. EU leaders will take a final decision in December. \\
\textcolor{blue}{GBDA w/ fluency} & \textcolor{blue}{Business (100\%)} & Turkey a step closer to Brussels The \textcolor{blue}{eurozone Union} is set to give the green light later today to \\ & & accession talks with \textcolor{blue}{Barcelona}. EU leaders will take a final decision in December. \\
\textcolor{red}{GBDA w/o fluency} & \textcolor{red}{Business (77\%)} &  Turkey a step closer to \textcolor{red}{Uber} \textcolor{red}{Thecom} Commission is set to give the green light later today to \\ & & \textcolor{red}{accessrage negotiations} with Turkey. EU leaders will take a final decision in December. \\[0.5em]

Original    & Science (76\%) &  Worldwide PC Market Seen Doubling by 2010  NEW YORK (Reuters) - The number of personal \\ & & computers worldwide is expected to double to about 1.3 billion by 2010,  driven by explosive growth \\ & & in emerging markets such as China,  Russia and India, according to a report released on Tuesday by \\ & &  Forrester Research Inc. \\
\textcolor{blue}{GBDA w/ fluency} & \textcolor{blue}{Business (98\%)} &  Worldwide PC \textcolor{blue}{Index} Seen Doubling by 2010  NEW YORK (Reuters) - The number of personal \\ & &  \textcolor{blue}{consumers} worldwide is expected to double to about 1.3 billion by 2010,  driven by explosive growth \\ & & in emerging markets such as China,  Russia and India, according to a report released on Tuesday by \\ & &  Forrester Research Inc. \\
\textcolor{red}{GBDA w/o fluency} & \textcolor{red}{Business (96\%)} &  Worldwide PC Market Seen Doubling by \textcolor{red}{2010qua} NEW YORK ( \textcolor{red}{REUTERSrow} - The number of \\ & &  personal computers worldwide \textcolor{red}{pensions} expected to \textcolor{red}{doublearound} about 1.3 billion \textcolor{red}{audits investors},  \\ & &  driven by explosive growth in emerging markets such as \textcolor{red}{Chinalo ru} Russia and \textcolor{red}{Yug Holo} \\ & &  according to a report released \textcolor{red}{onTue} by  Forrester Research Inc. \\[0.5em]
    \end{tabular}
    }
    \caption{\label{tab:fluency_constraint}Examples of adversarial texts on AG News generated with and without the fluency constraint. Without the fluency constraint, the constructed adversarial text tends to contain more nonsensical token combinations.}
\end{table*}

%% file: sections/conclusion.tex
\section{Conclusion and Future Work}
\label{sec:conclusion}

We presented GBDA, a framework for gradient-based white-box attacks against text transformers. The main insight in our method is the search formulation of a parameterized adversarial distribution rather than of a single adversarial examples, which enables efficient gradient-based optimization and differentiable constraints such as perplexity and BERTScore. Our attack is highly effective on a variety of natural language tasks and model architectures, and attains state-of-the-art attack performance in both the white-box and black-box transfer attack settings.

\paragraph{Limitations.} One clear limitation of our method is the restriction to only token replacements. Indeed, our adversarial distribution formulation using the Gumbel-softmax does not trivially extend to token insertions and deletions. This limitation may adversely affect the naturalness of the generated adversarial examples. We hope that our adversarial distribution framework can be extended to incorporate a broader set of token-level changes.

In addition, the adversarial distribution $P_\Theta$ is highly over-parameterized. Despite most adversarial examples requiring only a few token changes, the distribution parameter $\Theta$ is of size $n \times V$, which is especially excessive for longer sentences. Future work may be able to reduce the number of parameters without affecting the optimality of generated adversarial examples.